\documentclass{article}
\usepackage{microtype}
\usepackage{graphicx}
\usepackage{subfigure}
\usepackage{booktabs} 
\usepackage{nicefrac}       
\usepackage{amsmath}
\usepackage{amssymb}
\usepackage{mathbbol}
\usepackage{indentfirst}
\usepackage{enumitem}
\setenumerate[1]{itemsep=0pt,partopsep=0pt,parsep=\parskip,topsep=5pt} \setitemize[1]{itemsep=-6pt,partopsep=-6pt,parsep=\parskip,topsep=-4pt} \setdescription{itemsep=0pt,partopsep=0pt,parsep=\parskip,topsep=5pt}

\usepackage{amsthm}
\usepackage{wrapfig}
\usepackage{stmaryrd}
\usepackage{multirow}
\usepackage{array}
\makeatletter
\newcommand{\thickhline}{%
	\noalign {\ifnum 0=`}\fi \hrule height 1pt
	\futurelet \reserved@a \@xhline
}
\newcolumntype{"}{@{\hskip\tabcolsep\vrule width 1pt\hskip\tabcolsep}}
\makeatother

\newtheorem{thm}{Theorem}
\newtheorem{lem}{Lemma}
\newtheorem{prop}{Proposition}

\newtheorem{obv}{Observation}

\usepackage{hyperref}
\usepackage{bbm}

\usepackage[accepted]{icml2019}

\icmltitlerunning{Multivariate-Information Adversarial Ensemble}

\begin{document}
	
	\twocolumn[
	\icmltitle{Multivariate-Information Adversarial Ensemble for Scalable\\ Joint Distribution Matching}
	
	
	
	\icmlsetsymbol{equal}{*}
	
	\begin{icmlauthorlist}
		\icmlauthor{Ziliang Chen}{equal,sysu}
		\icmlauthor{Zhanfu Yang}{equal,purdue}
		\icmlauthor{Xiaoxi Wang}{equal,sysu}
		\icmlauthor{Xiaodan Liang}{sysu}
		\icmlauthor{Xiaopeng Yan}{sysu}
		\icmlauthor{Guanbin Li}{sysu}
		\icmlauthor{Liang Lin}{sysu}
	\end{icmlauthorlist}
	
	\icmlaffiliation{sysu}{Sun Yat-sen University, China}
	\icmlaffiliation{purdue}{Purdue University, USA}
	
	\icmlcorrespondingauthor{Liang Lin}{linliang@ieee.org}
	
	\icmlkeywords{Adversarial Machine Learning, ICML}
	
	\vskip 0.3in
	]
	
	
	
	\printAffiliationsAndNotice{\icmlEqualContribution} 
	
	\begin{abstract}
		A broad range of cross-$m$-domain generation researches boil down to matching a joint distribution by deep generative models (DGMs). Hitherto algorithms excel in pairwise domains while as $m$ increases, remain struggling to scale themselves to fit a joint distribution. In this paper, we propose a domain-scalable DGM, \emph{i.e.}, MMI-ALI for $m$-domain joint distribution matching. As an $m$-domain ensemble model of ALIs \cite{dumoulin2016adversarially}, MMI-ALI is adversarially trained with maximizing \emph{Multivariate Mutual Information} (MMI) \emph{w.r.t.} joint variables of each pair of domains and their shared feature. The negative MMIs are upper bounded by a series of feasible losses that provably lead to matching $m$-domain joint distributions. MMI-ALI linearly scales as $m$ increases and thus, strikes a right balance between efficacy and scalability. We evaluate MMI-ALI in diverse challenging $m$-domain scenarios and verify its superiority.  
	\end{abstract}\vspace{-10pt}
	\section{Introduction}
	\label{submission}
	\vspace{-6pt}Remarkable advances of Deep Generative Models (DGMs), \emph{e.g.}, \emph{Generative Adversarial Net} (GAN) \cite{goodfellow2014generative}, give rise to a variety of cross-domain generation and transfer tasks, \emph{e.g.}, label-to-image translation \cite{isola2017image,Wang2018Video}, visual / text style transfers \cite{shen2017style,zhu2017unpaired}, \emph{etc}. In these scenarios, examples drawn from one domain transform their appearances via DGMs to synthesize the data patterns that belong to the other domains. This magic is formally interpreted as learning a joint distribution \emph{w.r.t.} multi-domain random variables. Specifically, suppose that $m$ ($\forall m\hspace{-0.2em}\in\mathbb{N}_+\hspace{-0.1em}$) domains underly marginal distributions $\{p_1,\cdots,p_m\}$. Given an example $\boldsymbol{x}_i\sim p_i$ ($\forall i\in[m]=\{1,\cdots,m\}$), DGMs generate $\boldsymbol{x}_j$ ($\forall j\in[m]$,$j\neq i$) to satisfy the equation:
	\vspace{-6pt}\begin{equation}\begin{aligned}
	p(\boldsymbol{x}_1,\cdots,\boldsymbol{x}_m) :&= p(\{\boldsymbol{x}_j\}_{j\in[m]\&j\neq i}|\boldsymbol{x}_i)p(\boldsymbol{x}_i)
	\\&=p_{\Theta}(\{\boldsymbol{x}_j\}_{j\in[m]\&j\neq i}|\boldsymbol{x}_i)p(\boldsymbol{x}_i)\label{goal} 
	\end{aligned}
	\end{equation}where $p(\boldsymbol{x}_1,\cdots,\boldsymbol{x}_m)$ denotes the joint distribution on $m$-domain random variables. $p(\{\boldsymbol{x}_j\}_{j\in[m]\&j\neq i}|\boldsymbol{x}_i)$ is the conditional distribution \emph{w.r.t.} $\boldsymbol{x}_i$, and $p_{\Theta}(\{\boldsymbol{x}_j\}_{j\in[m]\&j\neq i}|\boldsymbol{x}_i)$ is parametrized from DGMs to match the $m$-domain joint distribution ($\Theta$ indicates the parameters of those DGMs). Eq.\ref{goal} is connected with a broad set of GAN-based DGMs. Particularly when $m=2$, (\ref{goal}) refers to finding a pair of generation nets to model $p(\boldsymbol{x}_2|\boldsymbol{x}_1)$ and 
	$p(\boldsymbol{x}_1|\boldsymbol{x}_2)$, exactly the learning goal shared by c-GAN \cite{isola2017image}, CycleGAN \cite{zhu2017unpaired,kim2017learning,yi2017dualgan} and other DGM methods \cite{dumoulin2016adversarially,li2017alice}. 
	
	Despite rapid progresses in learning paired-domain joint distribution, existing DGMs seldom prepare for the challenges as $m\hspace{-0.4em}>\hspace{-0.4em}2$, notably, the balance between model efficacy and scalability. On one hand, to cover $m(m-1)$ cross-domain transfer cases, most DGMs, \emph{e.g.}, CycleGAN and JointGAN \cite{Pu2018JointGAN}, have to deploy the same amount of (or even more) generation nets to learn $m$-domain joint distributions. It lacks efficiency in parameters and in turn, hinders them to capture richer information to improve their performances. On the other hand, recent heuristic methods, \emph{i.e.}, StarGAN \cite{choi2017stargan}, attempt to suit all the transfer tasks by a single pipeline where each domain is treated as a class. Their pipelines are indeed scalable but the algorithms do not promise them to learn joint distributions. In fact, this line of methods can be technically fragile: If the supports of $\{p_i\}^m_{i=1}$ tend to intersect, treating domains as classes will fail and arouse serious model collapse.
	
	In this paper, we focus on matching a $m$-domain joint distribution in a scalable and effective way. Instead of hacking a complex DGM pipeline, we revisit a famous \emph{Adversarially Learned Inference} (ALI) \cite{dumoulin2016adversarially} model from a prospective of \hspace{-0.2em} ensemble \cite{Polikar2009Ensemble}.\hspace{-0.3em} We assign $m$ ALIs (allowed to share some of parameters) to each domain for learning $m$ domain marginals by sharing their feature variables. By this mutual feature variable, each sample from domain $i$ can be encoded to a feature by the inference net in the $i^{th}$ ALI, then mapped into the $j^{th}$ domain ($j\neq i$) by the generation net in the $j^{th}$ ALI. This $m$ inference-generation ensemble enable $m(m-1)$ transfer cases and more importantly, may lead to $m$-domain joint distribution by appropriately regulating cross-domain dependency. 
	
	Specifically, we reframe this $m$-ALI ensemble trained with maximizing \emph{multivariate mutual information} (MMI) \cite{bell2003co,Mcgill2003Multivariate}. 
	The MMIs act on arbitrary joint variables originating from each pair of domains and the domain-shared feature, which implies that $m$-domain information flow may exchange via their mutual feature. This observation nails down to a series of upper bounds that indicates conditional generation \cite{isola2017image} and cycle consistency \cite{zhu2017unpaired}. They are provably connected with matching a $m$-domain joint distribution and make the $m$-ALI ensemble our final model, \emph{i.e.}, MMI-ALI.
	
	\vspace{-4pt}
	MMI-ALI mainly contributes as:\vspace{-4pt}

	\textbf{1).} MMI-ALI is linearly-scalable with $m$ and more importantly, holds a series of loss upper bounds for provable joint distribution matching. 
	\vspace{-6pt}
	
	\textbf{2).} MMI-ALI revisit classical ALI from a view of ensemble model and learn with a adversarial ensemble loss (Sect.\ref{2.4}), which are powerful for cross-domain generative modeling 
	\vspace{-6pt}
	
	\textbf{3).} A variety of $m$-domain experiments ($m\geq2$) are placed in diverse scenarios, \emph{e.g.}, $6$-domain setup, visual / text style transfer, \emph{etc}. The evaluation in supervised and unsupervised learning demonstrate the superiority of MMI-ALI. 
	\vspace{-4pt}
	
	\textbf{Related work.}
	Joint distribution matching has been considerably discussed in pairwise domain setups. Relevant researches based on GANs are classed into two lines. Models in the first line present as bidirectional DGMs associated with sample generation and feature inference, \cite{dumoulin2016adversarially,donahue2016adversarial,Tolstikhin2017Wasserstein,belghazi2018hierarchical}, real-real domain translations, \emph{e.g.}, CycleGANs \cite{zhu2017unpaired,kim2017learning,yi2017dualgan}, the variants \cite{hoffman2017cycada,gan2017triangle} and other adversarial dual learning models \cite{Ulyanov2017It,deng2017structured}. When cross-real-domain data are given in pairs, the second branch is connected with c-GAN \cite{isola2017image} and other conditional adversarial DGMs \cite{reed2016learning,reed2016generative,pathak2016context,Wang2018Video} . \cite{li2017alice} shows their relationships by conditional entropy (CE). Our paper extends it into $m$-domain scenarios.    
	
	\vspace{-4pt}In $m$-domain setup, joint distribution becomes more cumbersome to learn and a few of recent DGMs refer to this problem. To the best of our knowledge, JointGAN \cite{Pu2018JointGAN} is the only existing research that promises (\ref{goal}) when $m>2$. JointGAN chases for fully learning joint distribution, but ignores the scalability when $m$ increases and requires $C^3_{m}$ generative modules to attain $m(m-1)$ cross-domain transformations. StarGAN \cite{choi2017stargan} and its variants \cite{Zhao2018Modular,Kameoka2018StarGAN} use a domain-shared backbone where each domain is viewed as a class. They cast $m$-domain transfer to a category generation problem and do not aim to learn a joint distribution. 

\vspace{-8pt}\section{Multivariate Mutual Information Adversarially Learned Inference}\vspace{-6pt}
In this section, we elaborate MMI-ALI in the following routine: \textbf{1).} We introduce ALI (Sect.\ref{2.1}) and how it leads to an ensemble to achieve $m\hspace{-0.1em}(m\hspace{-0.1em}-\hspace{-0.1em}1\hspace{-0.1em})$ cross-domain transfer tasks (Sect.\ref{2.2}); \textbf{2).} We show the limitation of the $m$-ALI ensemble in cross-domain transfer (Sect.\ref{mov}) and how MMI induces a feasible regulation for the $m$-ALI ensemble to learn a joint distribution (Sect.\ref{2.3}). \textbf{3).} We provide the adversarial ensembel learning algorithm of MMI-ALI (Sect.\ref{2.4}).  
All proofs are deferred in our Appendix.A.
\begin{figure}[t]\centering
	\includegraphics[width=3.2in]{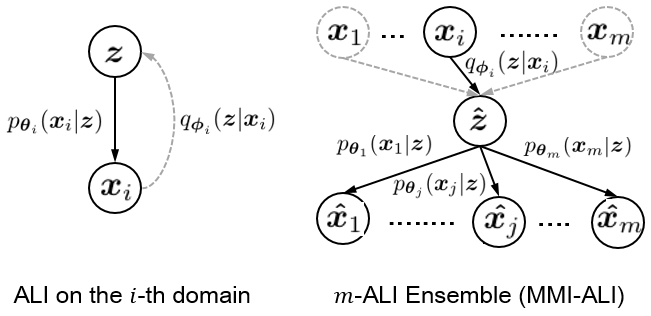}
	\vspace{-10pt}\caption{
		The overviews of ALI and $m$-ALI ensemble. MMI-ALI is learned from $m$-ALI ensemble with MMI constraints (Sect.\ref{2.3}).
	}\label{mmiali}\vspace{-14pt}
\end{figure}
\vspace{-10pt}
\subsection{Preliminary: Adversarially Learned Inference}\label{2.1}\vspace{-6pt}
ALI is a bidirectional DGM derived from GAN, as it additionally incorporates an inference net trained with a generation net by playing against a discriminator. More specifically, in our context, suppose that a ALI model refers to generating a fake domain-$i$ example $\boldsymbol{\hat{x}}_i$ ($\forall i\in[m]$). Without loss of generality, we employ a distribution $q(\boldsymbol{z})$ as a prior on feature space $\mathbb{R}^{d}$, e.g. $q(\boldsymbol{z})=\mathcal{N}(\mathbf{0}^d,\mathbf{I}^{d\times d})$. Under the nonparametric assumption, we present the generation and inference nets by conditional distributions $p_{\boldsymbol{\theta}_i}(\boldsymbol{\hat{x}}_i|\boldsymbol{z})$ and $q_{\boldsymbol{\phi}_i}(\boldsymbol{\hat{z}}|\boldsymbol{x}_i)$, where $\boldsymbol{\theta}_i$, $\boldsymbol{\phi}_i$ denote their parameters and their inputs $\boldsymbol{z}$, $\boldsymbol{x}_i$ are treated as the conditions. In this manner, ALI casts an adversarial game between $p_{\boldsymbol{\theta}_i}$, $q_{\boldsymbol{\phi}_i}$ and a $\boldsymbol{\omega}_i$-parameterized critic net (discriminator) $f_{\boldsymbol{\boldsymbol{\omega}}_i}$ in
\begin{equation}\label{ali}
\begin{aligned}
\underset{\boldsymbol{\boldsymbol{\theta}_i},\boldsymbol{\phi}_i}{\min} \ \underset{\boldsymbol{\boldsymbol{\omega}_i}}{\max} \ \mathcal{L}^{(i)}_{\rm ALI}(\boldsymbol{\boldsymbol{\theta}}_i, \boldsymbol{\phi}_i, \boldsymbol{\omega}_i) = \ \ \ &\\ \mathbb{E}_{\boldsymbol{x}_i\sim p(\boldsymbol{x}_i),\hat{\boldsymbol{z}}\sim q_{\boldsymbol{\boldsymbol{\phi}}_i}(\hat{\boldsymbol{z}}|\boldsymbol{x}_i)}\big[\log f_{\boldsymbol{\omega}_i}(&\boldsymbol{x}_i,\boldsymbol{\hat{z}})\big]\\
+ \ \mathbb{E}_{\boldsymbol{\hat{x}}_i\sim p_{\boldsymbol{\theta}_i}(\boldsymbol{\hat{x}}_i|\boldsymbol{z}),z\sim q(\boldsymbol{z})}&[\log\big(1-f_{\boldsymbol{\omega}_i}(\boldsymbol{\hat{x}}_i,\boldsymbol{z})\big)]
\end{aligned}
\end{equation}where $(\boldsymbol{x}_i,\hat{\boldsymbol{z}})$ denotes a real domain-$i$ example $\boldsymbol{x}_i$ with its corresponding feature $\hat{\boldsymbol{z}}$ inferred by $q_{\boldsymbol{\boldsymbol{\phi}}_i}$ and $(\hat{\boldsymbol{x}}_i,\boldsymbol{z})$ denotes a fake domain-$i$ sample $\hat{\boldsymbol{x}}_i$ generated from $\boldsymbol{z}\sim q(\boldsymbol{z})$ via $p_{\boldsymbol{\boldsymbol{\theta}}_i}$. $f_{\boldsymbol{\boldsymbol{\omega}}_i}(\cdot,\cdot)$ is a binary classifier that distinguishes each sample-feature joint pair drawn from either $q_{\boldsymbol{\boldsymbol{\phi}}_i}(\boldsymbol{x}_i,\hat{\boldsymbol{z}})$ or $p_{\boldsymbol{\theta}_i}(\boldsymbol{\hat{x}}_i,\boldsymbol{z})$. The minimax objective (\ref{ali}) encourages the iterative update between $\boldsymbol{\omega}_i$ and $\boldsymbol{\theta}_i$, $\boldsymbol{\phi}_i$. Similar to GAN, their resulting saddle point promises marginal matching on $p(\boldsymbol{x}_i)$, $q(\boldsymbol{z})$. 
\begin{lem}[\cite{dumoulin2016adversarially}]\label{lem2.1}
	The optimal generation, inference and critic nets w.r.t.,$\{\boldsymbol{\theta}_i^\ast,\boldsymbol{\phi}_i^\ast,\boldsymbol{\omega}_i^\ast\}$ $(\forall i\in[m])$ refer to a saddle point in Eq.\ref{ali} $\iff$ $ p_{\boldsymbol{\theta}_i^\ast}(\boldsymbol{x}_i|\boldsymbol{z})q(\boldsymbol{z})=q_{\boldsymbol{\phi}_i^\ast}(\boldsymbol{z}|\boldsymbol{x}_i)p_i(\boldsymbol{x}_i)$.
\end{lem}

\vspace{-8pt}\subsection{$m$-ALI Ensemble}\label{2.2}
With regards to $m$ domains, there can be $m$ ALIs that share the feature variable $\boldsymbol{z}$ to make marginal matchings on their own. It inspires an ensemble that associates $m$ domains to enable $m(m-1)$ cross-domain data transformations. As illustrated in Fig.\ref{mmiali}.Right, suppose that $\forall \boldsymbol{x}_i\sim p_{i}$ is demanded to transform to the other $j^{th}$ domain ($\forall i,j\in[m]$, $j\neq i$). By the aid of inference net $q_{\boldsymbol{\phi}_i}$ in the $i^{th}$ ALI, it is able to encode $\boldsymbol{x}_i$ into a domain-agnostic feature $\boldsymbol{\hat{z}}$, and then use the generation net $p_{\boldsymbol{\theta}_j}$ in the $j^{th}$ ALI to decode $\boldsymbol{\hat{z}}$ into $\boldsymbol{\hat{x}}_j$. This cross-domain generative process can be formulated as:
\vspace{-6pt}\begin{equation}
\begin{aligned}
p_{\boldsymbol{\Phi},\boldsymbol{\Theta}}&(\{\boldsymbol{\hat{x}}_j\}_{j\in[m]\&j\neq i}|\boldsymbol{x}_i)\\ =&\int p_{\boldsymbol{\Phi},\boldsymbol{\Theta}}(\{\boldsymbol{\hat{x}}_j\}_{j\in[m]\&j\neq i}|\boldsymbol{\hat{z}},\boldsymbol{x}_i) p_{\boldsymbol{\Phi},\boldsymbol{\Theta}}(\boldsymbol{\hat{z}}|\boldsymbol{x}_i)d\boldsymbol{\hat{z}}\\
=&\int \underbrace{\Big(\prod_{j\in[m]\&j\neq i}p_{\boldsymbol{\Phi},\boldsymbol{\Theta}}(\boldsymbol{\hat{x}}_j|\boldsymbol{\hat{z}})\Big)}_{{\mbox{\tiny$\begin{array}{c}{\rm Given \ } \boldsymbol{\hat{z}}, \ \{\boldsymbol{\hat{x}}_j\}_{j\in[m]\&j\neq i} {\ \rm and \ } \\
			\boldsymbol{x}_i \ { \rm are \ independent}\end{array}$}} } p_{\boldsymbol{\Phi},\boldsymbol{\Theta}}(\boldsymbol{\hat{z}}|\boldsymbol{x}_i)d\boldsymbol{z}\\=\int &\prod_{j\in[m], j\neq i} p_{\boldsymbol{\theta}_j}(\hat{\boldsymbol{x}}_j|\boldsymbol{\hat{z}})q_{\boldsymbol{\phi}_i}(\boldsymbol{\hat{z}}|\boldsymbol{x}_i)d\boldsymbol{\hat{z}}, \ s.t. \forall i\in[m]\label{m2}				
\end{aligned}
\end{equation}where we summarize the parameters of $m$-domain generation, inference, critic nets by $\boldsymbol{\Phi}=\{\boldsymbol{\phi}_i\}^m_{i=1}$, $\boldsymbol{\Theta}=\{\boldsymbol{\theta}_i\}^m_{i=1}$, $\boldsymbol{\Omega}=\{\boldsymbol{\omega}_i\}^m_{i=1}$.
As a cross-$m$-domain generative model, the $m$-ALI ensemble in (\ref{m2}) presents two advantages. 
\begin{itemize}\vspace{-1pt}
	\item \textbf{Scalability}: (\ref{m2}) is linearly-scalable with $m$. For sub-nets $\{q_{\boldsymbol{\phi}_i}\}^m_{i=1}$ and $\{p_{\boldsymbol{\theta}_i}\}^m_{i=1}$, it is possible to share their high-level layers across domains, as $m$-domain ALIs share their  feature variable $\boldsymbol{z}$.  
	\vspace{-0pt}
	\item \textbf{Generative model capability}: \hspace{-0.5em} According to Lemma.\ref{lem2.1}, (\ref{m2}) with $\boldsymbol{\phi}_i^\ast$ and $\boldsymbol{\theta}_j^\ast$ promises the transformed item $\boldsymbol{\hat{x}}_j$ following the true domain marginal $p_{j}$:
\end{itemize}
\begin{prop}
	Given a pair of domains $\forall i,j\in[m]$, $i\neq j$, their well-trained ALIs (in Lemma.1) construct a cross-domain transfer process $p_{\boldsymbol{\Phi},\boldsymbol{\Theta}}(\boldsymbol{\hat{\boldsymbol{x}}}_j|\boldsymbol{x}_i)$ that satisfies
	\begin{displaymath}
	p_{\boldsymbol{\Phi}^\ast,\boldsymbol{\Theta}^\ast}(\boldsymbol{\hat{\boldsymbol{x}}}_j)=\int p_{\boldsymbol{\Phi}^\ast,\boldsymbol{\Theta}^\ast}(\boldsymbol{\hat{\boldsymbol{x}}}_j|\boldsymbol{x}_i)p_i(\boldsymbol{x}_i)d\boldsymbol{x}_i=p_{j}(\boldsymbol{\hat{x}}_j)
	\end{displaymath}
\end{prop}\label{prop1} where $p_{\boldsymbol{\Phi},\boldsymbol{\Theta}}(\boldsymbol{\hat{\boldsymbol{x}}}_j|\boldsymbol{x}_i)$ is the parameterized marginal of (\ref{m2}).
\subsection{MMI-ALI: Motivation}\label{mov}

\textbf{How to learn $m$-ALI ensemble.} As we previously discuss, $m$-ALI ensemble is a promising non-parametric model to achieve $m(m-1)$ cross-domain transfer, as the scalability and generative model capability have verified its potential. But the vital problem is, how to encourage the $m$-ALI ensemble to learn a $m$-domain joint distribution. Obvisouly, since each ALI model in $m$-ALI ensemble is independently trained, no cross-domain dependencies enforce $p_{\boldsymbol{\Phi},\boldsymbol{\Theta}}$ to approximate the joint distribution $p(\boldsymbol{x}_1,\cdots,\boldsymbol{x}_m)$. As long as generated data can match domain marginals (Proposition.1), (\ref{m2}) may tolerate all erratic cross-domain transfer. To tackle this problem, we first need to understand how to match a joint distribution in the $m$-domain scenario.  

\vspace{-4pt}
\textbf{Criterion for $m$-domain joint distribution matching.}
In terms of supervised and unsupervised learning, joint distribution matching presents as satisfying different criterion.
\textbf{1).} In \emph{supervised learning}, we have access to draw samples from the true joint density $p(\boldsymbol{x}_1,\cdots,\boldsymbol{x}_m)$ and each of them presents as a $m$-tuple. Hence $p(\{\boldsymbol{x}_i\}^m_{i=1})$ can be learned by minimizing the log-likelihood estimator: 
\begin{equation}
{\min}_{\boldsymbol{\Phi},\boldsymbol{\Theta}} \ -\mathbb{E}_{p}\big[\log p_{\boldsymbol{\Phi},\boldsymbol{\Theta}}(\{\boldsymbol{x}_i\}^m_{i=1})\big]\label{c1}\end{equation}
\textbf{2).} In \emph{unsupervised learning}, data across domains are unparalleledly aligned so that no access is provided to draw $m$-tuple from $p(\boldsymbol{x}_i,\cdots,\boldsymbol{x}_m)$. In the pairwise domain setup \cite{zhu2017unpaired}, the unsupervised learning is typically considered as a cross-domain data reproduction problem that decreasing their conditional entropy (CE) theoretically helps to solve (see more in \citealt{li2017alice}):
\begin{equation}
\begin{aligned}
{\min}_{\boldsymbol{\Phi},\boldsymbol{\Theta}} \ H\big(\boldsymbol{x}_i|\hat{\boldsymbol{x}}_j\big)=-\mathbb{E}_{p_{\boldsymbol{\Phi},\boldsymbol{\Theta}}}\big[\log p_{\boldsymbol{\Phi},\boldsymbol{\Theta}}(\boldsymbol{x}_i|\hat{\boldsymbol{x}}_j)\big]\label{ce}
\end{aligned}
\end{equation}where $H\big(\boldsymbol{x}_i|\hat{\boldsymbol{x}}_j\big)$ measures the input reproduction uncertainty \emph{w.r.t.} $\boldsymbol{x}_i$ in the condition of $\hat{\boldsymbol{x}}_j$, \emph{i.e.}, what the input has produced.  
In our scenario, we develop (\ref{ce}) to incorporate $m$-domain variables 
\begin{equation}\begin{aligned}
{\min}_{\boldsymbol{\Phi},\boldsymbol{\Theta}} \ H&\big(\boldsymbol{x}_i|\{\boldsymbol{\hat{x}}_j\}_{{j\in[m]\&j\neq i}}\big)\\=-\mathbb{E}_{p_{\boldsymbol{\Phi},\boldsymbol{\Theta}}}&\big[\log p_{\boldsymbol{\Phi},\boldsymbol{\Theta}}(\boldsymbol{x}_i|\{\boldsymbol{\hat{x}}_j\}_{j\in[m]\&j\neq i})\big]\label{c2}
\end{aligned}
\end{equation}where $\forall i\in[m]$, $\boldsymbol{x}_i$ denotes an empirical draw from $p_i$; $\{\boldsymbol{\hat{x}}_j\}^m_{j=1\&j\neq i}$ denote fake items generated from $\boldsymbol{x}_i$ via (\ref{m2}).

It is worth noting that, (\ref{c1}) (\ref{c2}) with $m\hspace{-0.2em}=\hspace{-0.2em}2$ refer to condition \cite{isola2017image} and cycle-consistency loss \cite{zhu2017unpaired} that have been widely-used in GAN-based DGM. But in general cases ($m\geq2$), they are typically intractable and disconnected with the learning algorithm of ALI. 

Rather than directly optimizing (\ref{c1}) (\ref{c2}), we prefer exploring the information-theoretic meaning behind $m$-domain joint distribution. In the next subsection, we introduce \emph{Multivariate Mutual Information} (MMI) and explain it in the $m$-ALI ensemble context. We derive feasible MMIs \emph{w.r.t.} each pair of domains and feature. They refer to a series of upper bounds that can also be interpreted as condition and cycle losses. They result in (\ref{c1}) (\ref{c2}) to promise $m$-ALI ensemble learn for joint distribution matching.

\begin{figure*}[t]\centering
	\includegraphics[width=6.3in]{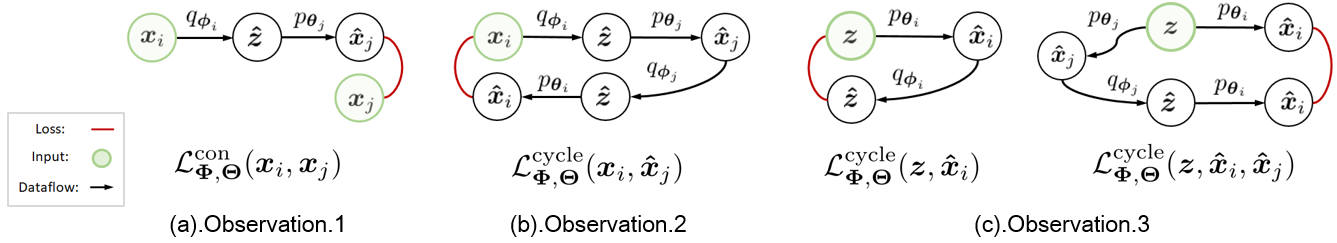}
	\vspace{-6pt}\caption{
		The diagram of constructing MMI-induced regularizations by generation and inference nets in $m$ ALIs. Best viewed in color.
	}\label{loss}\vspace{-16pt}
\end{figure*}

\subsection{MMI-Induced Regularization}\label{2.3}
Before diving into further technical analysis, let's quickly go through MMI, the pivotal ingredient of our regularization.

\textbf{Multivariate Mutual Information (MMI).} Given a pair of random variables $\boldsymbol{x}$, $\boldsymbol{y}$, \emph{Mutual Information} (MI) $I(\boldsymbol{x};\boldsymbol{y})$ quantifies the amount of information one of them contains about the other, \emph{i.e.},
\begin{equation}\begin{aligned}
I(\boldsymbol{x};\boldsymbol{y})=I(\boldsymbol{y};\boldsymbol{x}):=H(\boldsymbol{y})-H(\boldsymbol{y}|\boldsymbol{x})\end{aligned}
\end{equation}. Maximizing $I(\boldsymbol{x};\boldsymbol{y})$ relates to an invertible function that knowing one of $\boldsymbol{x}$, $\boldsymbol{y}$ almost reveals the other. MMI extends MI by including $n$ random variables $\boldsymbol{y}_1,\cdots,\boldsymbol{y}_n$ ($\forall n\in\mathbb{N}_+$). It can be recursively defined as 
\begin{equation}
\begin{aligned}
I(\boldsymbol{y}_1&;\cdots;\boldsymbol{y}_n)\\&:=I(\boldsymbol{y}_1;\cdots;\boldsymbol{y}_{n-1})-I(\boldsymbol{y}_1;\cdots;\boldsymbol{y}_{n-1}|\boldsymbol{y}_n)\label{mmi}
\end{aligned}
\end{equation}where $I(\boldsymbol{y}_1;\cdots;\boldsymbol{y}_{n-1}|\boldsymbol{y}_n)$ denotes \emph{Conditional Mutual Information} (CMI), the expectation of $I(\boldsymbol{y}_1;\cdots;\boldsymbol{y}_{n-1})$ when its value is conditioned on $\boldsymbol{y}_n$. 

\textbf{MMI for joint distribution matching.} MMI resembles the information-theoretic sense of MI. Maximizing $m$-domain MMI with respect to densities parameterized by $\boldsymbol{\Phi}$, $\boldsymbol{\Theta}$, \emph{i.e.}, $I_{\boldsymbol{\Phi},\boldsymbol{\Theta}}(\boldsymbol{x}_1;\cdots;\boldsymbol{x}_m)$, intuitively encourages discovering an identical information flow from one domain to the others. It corresponds to the cross-domain transfer $p_{\boldsymbol{\Phi},\boldsymbol{\Theta}}$ under $m$-domain joint distribution matching. However, on the basis of the recursive routine in (\ref{mmi}), $m$-variable MMI is comprised of $\mathcal{O}(2^m)$ entropy terms that can be positive or negative. It makes $I_{\boldsymbol{\Phi},\boldsymbol{\Theta}}(\boldsymbol{x}_1;\cdots;\boldsymbol{x}_m)$ intractable and formidable to extend with $m$. 
Besides, it probably arouses unstable optimization, as $I_{\boldsymbol{\Phi},\boldsymbol{\Theta}}(\boldsymbol{x}_1;\cdots;\boldsymbol{x}_m)$ may be unbounded.  

Instead of simultaneously considering $m$-domain variables, we tend to explore the linear combination of MMIs on each pair of domain variables $\boldsymbol{x}_i$, $\boldsymbol{x}_j$ with the $m$-domain-shared feature variable $\boldsymbol{z}$. In this principle, MMI $I_{\boldsymbol{\Phi},\boldsymbol{\Theta}}(\boldsymbol{x}_i;\boldsymbol{x}_j; \boldsymbol{z})$ has been covered $m(m-1)$ transfer cases and their maximizations are understood as
\vspace{-4pt}\begin{equation}
\begin{aligned}
\underset{\boldsymbol{\Phi},\boldsymbol{\Theta}}{\min} \hspace{1em}-\hspace{-1em}\sum_{i,j\in[m], i\neq j}\hspace{-1em}I_{\boldsymbol{\Phi},\boldsymbol{\Theta}}(\boldsymbol{x}_i;\boldsymbol{x}_j; \boldsymbol{z})
\end{aligned}\label{sb}
\end{equation}\vspace{-0pt}which implies the $m$-domain information flows exchange via their features. $I_{\boldsymbol{\Phi},\boldsymbol{\Theta}}(\boldsymbol{x}_i;\boldsymbol{x}_j; \boldsymbol{z})$ conceives two technical merits. First, three-variable MMI is always non-positive and thus, the minimization $-I_{\boldsymbol{\Phi},\boldsymbol{\Theta}}(\boldsymbol{x}_i;\boldsymbol{x}_j; \boldsymbol{z})$ is lower bounded by $0$, which substantially stabilizes the optimization process. Second, $-I_{\boldsymbol{\Phi},\boldsymbol{\Theta}}(\boldsymbol{x}_i;\boldsymbol{x}_j; \boldsymbol{z})$ can be pushed into a line of upper bounds that serve as condition and cycle-consistency losses. Their minimization results in (\ref{c1}) (\ref{c2}) that encourages $p_{\boldsymbol{\Phi},\boldsymbol{\Theta}}$ to learn the $m$-domain joint distribution. We are going to elaborate them.   

\textbf{Upper bounds.}  Derived from ALIs, $-I_{\boldsymbol{\Phi},\boldsymbol{\Theta}}(\boldsymbol{x}_i;\boldsymbol{x}_j; \boldsymbol{z})$ consists of generation and inference nets. Hence inputs underlie true distributions and may be drawn from either $m$ domain marginals $\{p_i\}^m_{i=1}$ or feature density $q(\boldsymbol{z})$. Suppose that $\boldsymbol{x}_i$, $\boldsymbol{x}_j$, $\boldsymbol{z}$ denote the observed variables \emph{w.r.t.} true distributions and $\boldsymbol{\hat{x}}_i$, $\boldsymbol{\hat{x}}_j$, $\boldsymbol{\hat{z}}$ denote the variables \emph{w.r.t.} $\boldsymbol{\Phi},\boldsymbol{\Theta}$-parameterized distributions. The upper bounds derived from $-I_{\boldsymbol{\Phi},\boldsymbol{\Theta}}(\boldsymbol{x}_i;\boldsymbol{x}_j; \boldsymbol{z})$ can be interpreted in three aspects. 

In the supervised case, training instances are $m$-tuples and for each domain-$i$ empirical draw, it is able to search its corresponding domain-$j$ empirical draw as the transformation groundtruth. In this scenario, $-I_{\boldsymbol{\Phi},\boldsymbol{\Theta}}(\boldsymbol{x}_i;\boldsymbol{x}_j; \boldsymbol{z})$ is bounded by the condition loss $\mathcal{L}^{\rm con}_{\boldsymbol{\Phi},\boldsymbol{\Theta}}(\boldsymbol{x}_i,\boldsymbol{x}_j)$ as below
\vspace{-1pt}\begin{obv}
	Given empirical draws from $p_i$ ($\forall i\in[m]$), in supervised learning, 
	\begin{equation}
	\begin{aligned}
	&-I_{\boldsymbol{\Phi},\boldsymbol{\Theta}}(\boldsymbol{x}_i;\boldsymbol{x}_j;\boldsymbol{\hat{z}})\leq H_{\boldsymbol{\Phi},\boldsymbol{\Theta}}(\boldsymbol{x}_i|\boldsymbol{x}_j)\\\leq&\hspace{-0.3em}\underset{{\boldsymbol{x}_i,\boldsymbol{x}_j\sim p_{i,j}}}{\mathbb{E}}\hspace{-1em}-\big[\log\int\hspace{-0.3em} p_{\boldsymbol{\theta}_i}(\boldsymbol{x}_i|\boldsymbol{\hat{z}})q_{\boldsymbol{\phi}_j}(\boldsymbol{\hat{z}}|\boldsymbol{x}_j)d\boldsymbol{\hat{z}}\big]\hspace{-0.2em}\triangleq\hspace{-0.2em} \mathcal{L}^{\rm con}_{\boldsymbol{\Phi},\boldsymbol{\Theta}}(\boldsymbol{x}_i,\boldsymbol{x}_j)
	\end{aligned}\label{con}
	\end{equation}where $p_{i,j}=p(\boldsymbol{x}_i,\boldsymbol{x}_j)$. \end{obv}In Fig.\ref{loss}.a., we show how to build $\mathcal{L}^{\rm con}_{\boldsymbol{\Phi},\boldsymbol{\Theta}}(\boldsymbol{x}_i,\boldsymbol{x}_j)$. The loss can be implemented by $l_1$/$l_2$ norms. 

In the unsupervised case, each empirical draw is separately given, therefore we have no access to $\boldsymbol{x}_j$. Distinct from (\ref{con}), the MMI turns into $I_{\boldsymbol{\Phi},\boldsymbol{\Theta}}(\boldsymbol{x}_i;\boldsymbol{\hat{x}}_j;\boldsymbol{\hat{z}})$ where $\boldsymbol{\hat{x}}_j$ implies that domain-$j$ samples are counterfeits and the bound constitutes a cross-domain cycle-consistency loss by means of $\boldsymbol{\hat{z}}$:
\begin{obv}
	Given empirical draws from $p_i$ ($\forall i\in[m]$), in unsupervised learning,
	\begin{equation}
	\begin{aligned}
	&-I_{\boldsymbol{\Phi},\boldsymbol{\Theta}}(\boldsymbol{x}_i;\boldsymbol{\hat{x}}_j;\boldsymbol{\hat{z}})\leq H_{\boldsymbol{\Phi},\boldsymbol{\Theta}}(\boldsymbol{x}_i|\boldsymbol{\hat{x}}_j)\\\leq&\underset{{\boldsymbol{x}_i,\boldsymbol{\hat{x}}_j\sim p_{\boldsymbol{\theta}_j,\boldsymbol{\phi}_i}}}{\mathbb{E}}\hspace{-1.5em}-\big[\log\int\hspace{-0.3em} p_{\boldsymbol{\theta}_i}(\boldsymbol{x}_i|\boldsymbol{\hat{z}})q_{\boldsymbol{\phi}_j}(\boldsymbol{\hat{z}}|\boldsymbol{x}_j)d\boldsymbol{\hat{z}}\big]\hspace{-0.2em}\triangleq\hspace{-0.2em}\mathcal{L}^{\rm cycle}_{\boldsymbol{\Phi},\boldsymbol{\Theta}}(\boldsymbol{x}_i,\boldsymbol{\hat{x}}_j)
	\end{aligned}
	\end{equation}where $p_{\boldsymbol{\theta}_j,\boldsymbol{\phi}_i}=p(\boldsymbol{x}_i)\int_{\boldsymbol{\hat{z}}}p_{\boldsymbol{\theta}_j}(\hat{\boldsymbol{x}}_j|\boldsymbol{\hat{z}})q_{\boldsymbol{\phi}_i}(\boldsymbol{\hat{z}}|\boldsymbol{x}_i)d\boldsymbol{\hat{z}}$.
\end{obv}$\mathcal{L}^{\rm cycle}_{\boldsymbol{\Phi},\boldsymbol{\Theta}}(\boldsymbol{x}_i,\boldsymbol{\hat{x}}_j)$ is constructed as illustrated in Fig.\ref{loss}.b.

The observations above presumed inputs drawn from the domain marginals $\{p_i\}^m_{i=1}$. If inputs are drawn from the feature distribution $q(\boldsymbol{z})$, $\boldsymbol{\hat{x}}_i$, $\boldsymbol{\hat{x}}_j$ would be generated from $\boldsymbol{z}$, and $-I_{\boldsymbol{\Phi},\boldsymbol{\Theta}}(\boldsymbol{\hat{x}}_i;\boldsymbol{\hat{x}}_j;\boldsymbol{z})$ is upper bounded by the conditional entropies $H_{\boldsymbol{\Phi},\boldsymbol{\Theta}}(\boldsymbol{z}|\boldsymbol{\hat{x}}_i)$ and $H_{\boldsymbol{\Phi},\boldsymbol{\Theta}}(\boldsymbol{\hat{x}}_j|\boldsymbol{\hat{x}}_i)$. They are equivalent to the cycle losses $\mathcal{L}^{\rm cycle}_{\boldsymbol{\Phi},\boldsymbol{\Theta}}(\boldsymbol{z},\boldsymbol{\hat{x}}_i)$ and $\mathcal{L}^{\rm cycle}_{\boldsymbol{\Phi},\boldsymbol{\Theta}}(\boldsymbol{z},\boldsymbol{\hat{x}}_i,\boldsymbol{\hat{x}}_j)
$, which are revealed in Fig.\ref{loss}.c.
\vspace{-3pt}\begin{obv}
	Given empirical draws from $q(\boldsymbol{z})$, \vspace{-5pt}\begin{equation}
	\begin{aligned}
	-I_{\boldsymbol{\Phi},\boldsymbol{\Theta}}(\boldsymbol{\hat{x}}_i;\boldsymbol{\hat{x}}_j;\boldsymbol{z})\leq H_{\boldsymbol{\Phi},\boldsymbol{\Theta}}(\boldsymbol{z}|\boldsymbol{\hat{x}}_i)+H_{\boldsymbol{\Phi},\boldsymbol{\Theta}}(\boldsymbol{\hat{x}}_j|\boldsymbol{\hat{x}}_i)
	\end{aligned}
	\end{equation}\vspace{-4pt}
	\begin{displaymath}
	\begin{aligned}
	H_{\boldsymbol{\Phi},\boldsymbol{\Theta}}(\boldsymbol{z}|\boldsymbol{\hat{x}}_i)&=\hspace{-1em}\underset{\boldsymbol{\hat{x}}_i\sim p_{\boldsymbol{\theta}_i}, \boldsymbol{z}\sim q(\boldsymbol{z})}{\mathbb{E}}\hspace{-2em}-\log q_{\boldsymbol{\phi}_i}(\boldsymbol{z}|\boldsymbol{\hat{x}}_i)\hspace{-0.2em}\triangleq\hspace{-0.2em}\mathcal{L}^{\rm cycle}_{\boldsymbol{\Phi},\boldsymbol{\Theta}}(\boldsymbol{z},\boldsymbol{\hat{x}}_i)\\
	H_{\boldsymbol{\Phi},\boldsymbol{\Theta}}(\boldsymbol{\hat{x}}_j|\boldsymbol{\hat{x}}_i)&=\hspace{-2em}\underset{{\mbox{\tiny$\begin{array}{c} \boldsymbol{z}\sim q(\boldsymbol{z}) \\
				\boldsymbol{\hat{x}}_i\sim p_{\theta_i},\hat{\boldsymbol{x}}_j\sim p_{\theta_j}   \end{array}$}}}{\mathbb{E}}\hspace{-2em}-\big[\log\int_{\boldsymbol{z}}\hspace{-0.3em}p_{\boldsymbol{\theta}_j}(\hat{\boldsymbol{x}}_j|\boldsymbol{z})q_{\boldsymbol{\phi}_i}(\boldsymbol{z}|\boldsymbol{\boldsymbol{\hat{x}}}_i)d\boldsymbol{z}\big]\\& \ \ \ \ \ \ \ \ \ \ \ \ \ \ \ \ \ \ \ \ \ \ \ \ \ \ \ \ \ \ \ \ \ \triangleq \mathcal{L}^{\rm cycle}_{\boldsymbol{\Phi},\boldsymbol{\Theta}}(\boldsymbol{z},\boldsymbol{\hat{x}}_i,\boldsymbol{\hat{x}}_j)
	\end{aligned}
	\end{displaymath}
\end{obv}\vspace{-8pt} 

Associate Observations (1-3) and we impose cross-domain structure dependencies on $\boldsymbol{\Phi}$, $\boldsymbol{\Theta}$ by
\begin{equation}
\begin{aligned}
\mathcal{R}_{\rm SL}(\boldsymbol{\Theta},\boldsymbol{\Phi})=\hspace{-1.2em}\sum_{i,j\in[m],i\neq j} \hspace{-1.2em} \mathcal{L}^{\rm con}_{\boldsymbol{\Phi},\boldsymbol{\Theta}}(&\boldsymbol{x}_i,\boldsymbol{x}_j)+\mathcal{L}^{\rm cycle}_{\boldsymbol{\Phi},\boldsymbol{\Theta}}(\boldsymbol{z},\boldsymbol{\hat{x}}_i)\\&+\mathcal{L}^{\rm cycle}_{\boldsymbol{\Phi},\boldsymbol{\Theta}}(\boldsymbol{z},\boldsymbol{\hat{x}}_i,\boldsymbol{\hat{x}}_j)\\
\mathcal{R}_{\rm UL}(\boldsymbol{\Theta},\boldsymbol{\Phi})=\hspace{-1.2em}\sum_{i,j\in[m],i\neq j} \hspace{-1.2em} \mathcal{L}^{\rm cycle}_{\boldsymbol{\Phi},\boldsymbol{\Theta}}(&\boldsymbol{x}_i,\boldsymbol{\hat{x}}_j)+\mathcal{L}^{\rm cycle}_{\boldsymbol{\Phi},\boldsymbol{\Theta}}(\boldsymbol{z},\boldsymbol{\hat{x}}_i)\\&+\mathcal{L}^{\rm cycle}_{\boldsymbol{\Phi},\boldsymbol{\Theta}}(\boldsymbol{z},\boldsymbol{\hat{x}}_i,\boldsymbol{\hat{x}}_j)
\end{aligned}\label{mmir}
\end{equation} where $\mathcal{R}_{\rm SL}$ / $\mathcal{R}_{\rm UL}$ respectively regulate the supervised / unsupervised learning and upper bound (\ref{sb}). It implies that the minimization of $\mathcal{R}_{\rm SL}$, $\mathcal{R}_{\rm UL}$ equal to maximizing the MMIs. By Proposition.1, desire that adversarial learning (\ref{ali}) encourages $\{p_i\}^m_{i=1}$ and parameterized domain marginals agree with a high likelihood to domain variables \big(\emph{i.e.}, $\boldsymbol{x}_i=\boldsymbol{\hat{x}}_i$ in (\ref{mmir})\big), then the minimization of $\mathcal{R}_{\rm SL}$, $\mathcal{R}_{\rm UL}$ leads to the joint distribution matching criterion (\ref{c1}),(\ref{c2}). 
\vspace{-2pt}\begin{thm}
	Suppose that true and parameterized domain marginal distributions maintain a high likelihood to domain variables, $\mathcal{R}_{\rm SL}\rightarrow 0$ leads to the optima in (\ref{c1}); $\mathcal{R}_{\rm UL}\rightarrow 0$ leads to the optima in (\ref{c2}).
\end{thm}\vspace{-8pt}
\subsection{Adversarial Ensemble Learning}\vspace{-4pt}\label{2.4}
Learning $m$-ALI ensemble by (\ref{mmir}) is able to capture the $m$-domain joint density. But it can be problematic as samples directly generated from $q(\boldsymbol{z})$ can be of low quality, \emph{e.g.}, due to the poorly-efficient sampling in a high-dimensional feature space. To overcome this issue, we invent a \emph{domain mixture adversarial ensemble (DMAE) loss} to refine (\ref{ali}) :
\begin{equation}
\begin{aligned}
\mathcal{L}^{(i)}_{\rm DMAE}(\boldsymbol{\Phi},\boldsymbol{\Theta},&\boldsymbol{\Omega})=\mathbb{E}_{\boldsymbol{x}_i,\hat{\boldsymbol{z}}\sim q_{\boldsymbol{\boldsymbol{\phi}}_i}(\boldsymbol{x}_i, \hat{\boldsymbol{z}})}\big[\log f_{\boldsymbol{\omega}_i}(\boldsymbol{x}_i,\boldsymbol{\hat{z}})\big]\\+\sum_{j=1}^{m}\pi_j &\Big(\mathbb{E}_{\boldsymbol{\hat{x}}_i\sim p_{\boldsymbol{\theta}_i}(\boldsymbol{\hat{x}}_i|\boldsymbol{z}),\boldsymbol{z}\sim q_{\boldsymbol{\phi}_j}}[\log\big(1-f_{\boldsymbol{\omega}_i}(\boldsymbol{\hat{x}}_i,\boldsymbol{z})\big)]\Big)
\end{aligned}\label{mix}
\end{equation}where $\sum_{j=1}^{m}\pi_j\hspace{-0.2em}=\hspace{-0.2em}1$ indicates the proportion of the domain mixture for adversary. Compared with (\ref{ali}) whose fake samples are solely generated from $q(\boldsymbol{z})$, $\mathcal{L}^{(i)}_{\rm DMAE}(\boldsymbol{\Phi},\boldsymbol{\Theta},\boldsymbol{\Omega})$ consider fake samples generated from the domain-encoded features, which are derived from the real samples that belong to the other domains, \emph{i.e.}, $\boldsymbol{z}\sim \int q_{\boldsymbol{\phi}_j}(\boldsymbol{z},\boldsymbol{x}_j)d\boldsymbol{x}_j$ ($\forall j\in[m]$). These fake samples converted from different domains are unified into the DMAE loss (\ref{mix}) to cheat the domain-$i$ critic net $f_{\boldsymbol{\omega}_i}$. It can be provably verified that, the adversarial ensemble learning retains the theoretical property of (\ref{ali}):
\vspace{-3pt}\begin{prop}\label{thm_g}
	The optimum of the generation, inference and critic networks in 
	\vspace{-6pt}\begin{equation}
	\underset{\boldsymbol{\boldsymbol{\Theta}},\boldsymbol{\Phi}}{\min} \ \underset{\boldsymbol{\boldsymbol{\Omega}}}{\max} \ (1-\gamma)\sum_{i=1}^{m}\mathcal{L}^{(i)}_{\rm ALI}+\gamma\sum_{i=1}^{m}\mathcal{L}^{(i)}_{\rm DMAE}\label{gen}
	\end{equation}refer to their saddle points in Lemma.\ref{lem2.1} if and only if $\forall i\in[m]$, there exist $ p_{\boldsymbol{\theta}_i^\ast}(\boldsymbol{x}|\boldsymbol{z})q(\boldsymbol{z})=q_{\boldsymbol{\phi}_i^\ast}(\boldsymbol{z}|\boldsymbol{x})p(\boldsymbol{x})$.
\end{prop}\vspace{-6pt} where $\gamma$ denotes the trade-off between (\ref{ali}) and DAME loss. Proposition.\ref{thm_g} demonstrates that, even if we change the learning objective (\ref{ali}), Lemma.\ref{lem2.1} and the other analysis based on (\ref{ali}) can be completely followed by the new objective (\ref{gen}).

Combining (\ref{mmir}) and (\ref{gen}), we formalize MMI-ALI as\begin{equation}\begin{aligned}
\underset{\boldsymbol{\boldsymbol{\Theta}},\boldsymbol{\Phi}}{\min} \ \underset{\boldsymbol{\boldsymbol{\Omega}}}{\max} &\  (1-\gamma)\sum_{i=1}^{m}\mathcal{L}^{(i)}_{\rm ALI}+\hspace{-0.2em}\gamma\sum_{i=1}^{m}\mathcal{L}^{(i)}_{\rm DMAE}+\hspace{-0.2em}\beta \ \mathcal{R}_{\rm SL}/\mathcal{R}_{\rm UL}
\end{aligned}
\end{equation}where $\mathcal{R}_{\rm SL}/\mathcal{R}_{\rm UL}$ are switched by supervised/unsupervised learning and $\beta>0$ denotes the loss-balance factor. 

\vspace{-6pt}\section{Experiments}\vspace{-6pt}
In this section, we propose diverse cross-$m$-domain experiments to evaluate our MMI-ALI in generative modeling and show the primal empirical results. More experiments (\emph{e.g.}, ablation) and visualization are founded in Appendix.B\footnote{\url{http://github.com/MintYiqingchen/MMI-ALI}.} 

\vspace{-6pt}\subsection{Balance between efficacy and scalability}
\begin{figure}[t]\centering
	\includegraphics[width=3.25in]{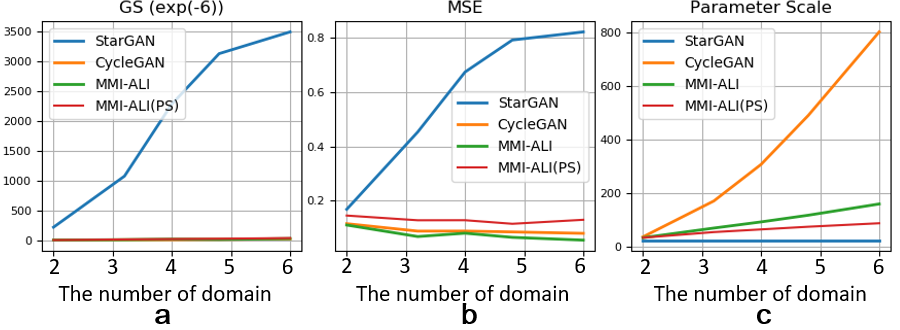}
	\vspace{-18pt}\caption{ Transfer evaluations with 2$\sim$6 synthetic domains: (a). Geometric Score (GS, lower is better); (b). Mean Square Error (MSE, lower is better); (c). Parameter Scale (lower is better). 
	}\vspace{-18pt}\label{sd}
\end{figure}

\vspace{-6pt}
Compared with existing methods, MMI-ALI strikes a right balance between model capacity and scalability. To highlight this merit, we design the first experiment on synthetic data domains with $m$ ranged in $2$$\sim$$6$. We choose $q(\boldsymbol{z})$ as an isotropic Gaussian $\mathcal{N}(\mathbf{0},\mathbf{I})$, then each density in $\{p_i\}^m_{i=1}$ is a 2D Gaussian Mixture Model (GMM) with 5 components $\mathcal{N}(\mathbf{0},0.2\mathbf{I})$. (As illustrated in Fig.\ref{sd}) Due to the simplicity of synthetic data, we only consider unsupervised learning across them. We evaluate MMI-ALI and its parameter-shared version termed ''MMI-ALI (PS)'', with CycleGAN and StarGAN. All of them are trained on $2048$ with vanilla GAN loss and tested on $1024$ examples drawn from each of $\{p_i\}^m_{i=1}$. For a fair comparison, all baselines use two-layered fully-connected nets with ReLU to generate data and make critics. $l_2$-norm is chosen as the cycle-consistency loss for all baseline during training.  

\textbf{Evaluation.} Two measures have been introduced. The first is \emph{geometric score} (GS) \cite{Khrulkov2018Geometry} that evaluates generation quality by comparing the topological properties of the supports behind the generated and true domain marginals. The other is \emph{mean squared error} (MSE) broadly used to measure the conditional density modeling via sample reconstruction quality across domains. Each baseline is performed in average of $m(m-1)$ transfer cases on two measures to thoroughly reflect the learned joint distribution. The results and parameters are shown in Fig.\ref{sd}.(a-b) and (c), respectively. Note that, StarGAN uses a domain-shared pipeline so that its parameter scale is almost consistent as $m$ increases. However, StarGAN's GS, MSE heavily suffer even in toy domains, due to its intrinsic vulnerability as we have discussed. Particularly, when there exists an overlap across domains, the examples drawn from the overlap (or close to the overlap) can belong to all of these domains. This phenomena is general (see our empirical results in real data) and StarGANs can do nothing to help. On the other hand, MMI-ALI and CycleGAN are close in GS and MSE, yet CycleGAN requires exponentially-increasing parameters. They demonstrate that MIM-ALIs remain convincing performances as they scale to the scenarios with more domains. 
We show more visualization results in SM. 

\begin{figure}[t]\centering
	\vspace{-3pt}\hspace{3em}\includegraphics[width=3.2in]{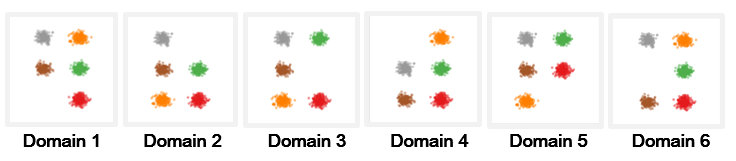}
	\vspace{-8pt}\caption{Synthetic domains used in our first experiments. As $m$ increases, they are proceedingly incorporated for multi-domain joint distribution leanring from left to right.}
	\vspace{-13pt}\label{6d}\label{Fig1}\vspace{2pt}
\end{figure} 

\vspace{-6pt}\subsection{Geometry-varying $m$ domains.}\vspace{-6pt}

Geometry-varying information is difficult to capture in generative modeling \cite{Sabour2017Dynamic}. Based on this challenge, our second experiment considers cross-$m$-domain generation where the $m$-domain samples present significant variation in geometry. We evaluate whether this information can be captured by MMI-ALI and the other baselines. 
\begin{figure}[t]\centering
	\includegraphics[width=3.2in]{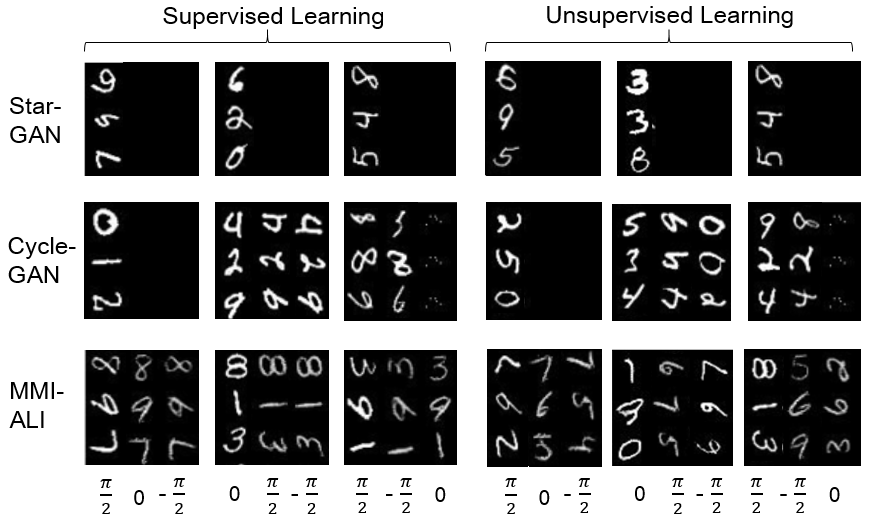}
	\vspace{-13pt}\caption{Cross-$3$-domain generation performed by StarGAN, CycleGAN and MMI-ALI (ours) in supervised and unsupervised learning setups. \emph{\textbf{For each sub-picture, the left column indicates inputs}} and the rest indicate the cross-domain transformed results. 
	}\label{digit}\vspace{-9pt}
\end{figure}

\begin{table}[h]\center
	\begin{footnotesize}
		\vspace{-6pt}\caption{ SSIM of StarGAN (ST), CycleGAN (CG) and MMI-ALI(MA) in supervised cross-domain generation case. }\label{t1}
		\begin{tabular}{lccccccccr}
			&$1\%$ &$5\%$ &$10\%$ \\
			\hline
			ST  &0.00 &0.00 &0.00 \\
			CG	& 0.32		&0.31		&0.35			\\
			MA	&\textbf{0.57}		&\textbf{0.68}		&\textbf{0.72}		
		\end{tabular}\vspace{-6pt}
	\end{footnotesize}
	\caption{ IS of StarGAN (ST), CycleGAN (CG) and MMI-ALI(MA) in unsupervised cross-domain generation case.}\label{t2}
	\begin{footnotesize}
		\begin{tabular}{lccccccccr}
			\thickhline
			&\begin{tiny}
				$-\frac{\pi}{2}\hspace{-0.3em}\rightarrow\hspace{-0.3em}0$
			\end{tiny}\hspace{-0.2em} &\begin{tiny}$\frac{\pi}{2}\hspace{-0.3em}\rightarrow\hspace{-0.3em}0$\end{tiny}\hspace{-0.2em} &\begin{tiny}$0\hspace{-0.3em}\rightarrow\hspace{-0.3em}\frac{\pi}{2}$\end{tiny}\hspace{-0.2em} &\begin{tiny}
			$-\frac{\pi}{2}\hspace{-0.3em}\rightarrow\hspace{-0.3em}\frac{\pi}{2}$
		\end{tiny}\hspace{-0.2em} &\begin{tiny}$-\frac{\pi}{2}\hspace{-0.3em}\rightarrow\hspace{-0.3em}0$\end{tiny}\hspace{-0.2em} &\begin{tiny}$\frac{\pi}{2}\hspace{-0.3em}\rightarrow\hspace{-0.3em}-\frac{\pi}{2}\hspace{-0.2em}$\end{tiny}\\
		ST\hspace{-0.4em}	& 1.00		&1.00		&1.00		& 1.00		&1.00		&1.00 			\\
		CG\hspace{-0.4em}	&8.34		&6.13		&2.25		&2.38	  &1.71	&1.04	\\
		Ours\hspace{-0.4em}&8.99		&9.01		&2.95		&3.86	  &3.31 &3.08\\
		\thickhline
	\end{tabular}
\end{footnotesize}\vspace{-18pt}
\end{table}
Specifically, we choose MNIST as the base domain, then rotate the images by $-\frac{\pi}{2}^\circ$, $\frac{\pi}{2}^\circ$ to create two other domains. Then MMI-ALI, CycleGAN and StarGAN are demanded to learn pattern transfer across the three domains in supervised and unsupervised learning setups. In supervised setup, data present as triplets so that each example from one domain has its corresponding groundturth in other domains. This information is not provided in unsupervised cases. In supervised case, we compare (supervised) MMI-ALI with CycleGAN and StarGAN augmented with condition loss used by c-GAN. In unsupervised case, we compare (unsupervised) MMI-ALI with ordinary CycleGAN and StarGAN. For a fair comparison, we standardize backbone behind the baselines in DCGAN \cite{dumoulin2016adversarially}, and they are trained with vanilla GAN and $l_1$-norm cycle losses.

\textbf{Evaluation.} In supervised learning setup, we measure transformed results by Structured SIMilarity (SSIM) \cite{Zhou2004Image}. The visualization and quantitative results are shown in Fig.\ref{sd} and Table.\ref{t2}, respectively. MMI-ALI is the \emph{only baseline} that can produce all transfer patterns. StarGAN collapses during training and create nothing for transfer. CycleGAN performs better than MMI-ALI in $0\rightarrow -\frac{\pi}{2},\frac{\pi}{2}$, however, fails in capturing larger rotation (\emph{e.g.}, $-\frac{\pi}{2} \rightarrow \frac{\pi}{2}$). It demonstrates a weakness of CycleGAN, which merely learns a pairwise joint distribution per time. In other word, it can not leverage $m$-domain knowledge to enhance the cross-domain generation performance. MMI-ALI avert this issue due to modeling $m$-domain joint distribution by ensemble. For more concrete evaluation, we provide different proportion of supervised data, \emph{i.e.}, $1\%$, $5\%$, $10\%$, to check how much the model can benefit from supervision. We find that in $3$-domain Rotated MNIST, cross-domain alignment can not significantly help StarGAN and CycleGAN to improve their joint distribution learning performance. But MMI-ALI can benefit from small amount of supervision.
\begin{figure}[t]\centering
	\includegraphics[width=3.2in]{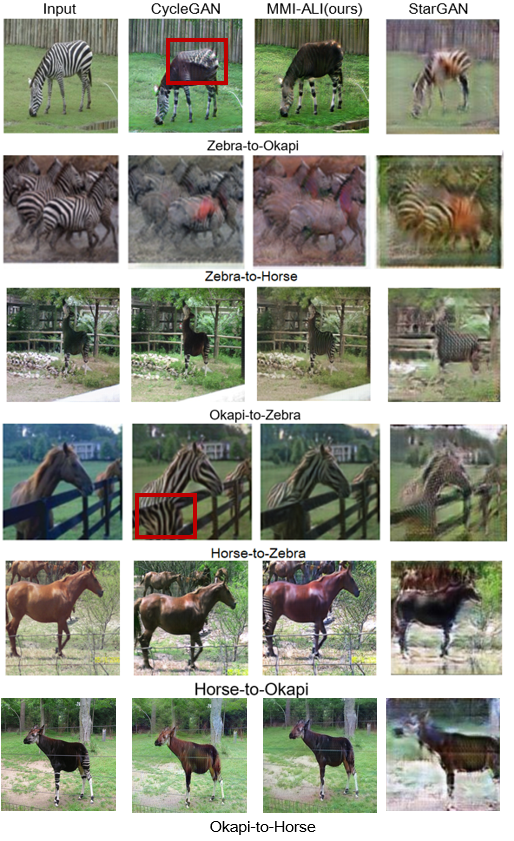}
	\vspace{-10pt}\caption{
		Style transfer on Zebra\&Horse\&Okapi. 
	}\label{vst}\vspace{-18pt}
\end{figure}
Cross-domain digit transformation conceives structure variation, thus, the patterns are difficult to capture without supervisions. This statement is verified in unsupervised results shown in Fig.\ref{sd}. Even so, our MMI-ALI is still powerful in generative modeling. To be specific, we evaluate the unsupervised generation by \emph{Inception Score} \cite{salimans2016improved}. MMI-ALI consistently outperform the other baselines across $6$ cross-domain generation scenarios.
\subsection{Cross-$m$-domain visual style transfer.}\vspace{-4pt}
In this experiment, we consider $3$-domain object transfiguration and $3$-heterogeneous-domain style transfer.		

In object transfiguration, evaluated DGMs are required to transform a specific part of an object to some target pattern whereas the other parts remain the same. One example is to translate a sort of animals (\emph{e.g.}, $1000$ classes in ImageNET ) to become another kind with visual similarity. In our experiment, we consider the $3$-object transfiguration in ${\rm Zebra}\leftrightarrow{\rm Horse}\leftrightarrow{\rm Okapi}$, where ${\rm Zebra}$ and ${\rm Horse}$ share their shapes while differ from the strip; then ${\rm Okapi}$ is \emph{``zebra-
	striped" on its legs with a ``horse-like" torso}. The experiment is conducted by reconfiguring the state-of-the-art residual-block-based \cite{He2015Deep} CycleGAN into MMI-ALI. For a fair comparison with CycleGAN, we depart the generator of CycleGAN as a pair of inference and generation net for our MMI-ALI, and follow the identical training tricks. Instead of using a non-informative prior, we apply $\boldsymbol{z} = \mu(\boldsymbol{z}) + \epsilon$ to provide features. As for StarGAN, we employ the official code reported in their original paper where their models are also built on ResNet.  

In $3$-heterogeneous-domain transfer, we consider Cityscape \cite{Cordts2016Cityscapes} as the base benchmark, then employ the real data and their segmentation labels to construct two domains (R and Seg). We further applied the pretrained sketch detector \cite{Xie2015Holistically} to generate the third domain (Ske). To this we are able to evaluate all baselines in unsupervised and supervised learning manners (Condition loss is used in the supervised case). We resemble the similar configuration and training strategy in object transfiguration. 


\textbf{Evaluation.} Amazon
Mechanical Turk (AMT) is employed to evaluate the object transfiguration experiment. We follow the perceptual evaluation from \cite{Dong2018Soft}, where workers are provided with a pair of generated image (ours and the other baseline), and given unlimited time to select the one more likely as a target domain image. In Cityscape, we take \emph{Frechet Inception Distance} (FID)\cite{Heusel2018GANs} 
and MSE as the metrics (MSE deferred in SM). 
\begin{table}[h]\centering
	\begin{footnotesize}
		\vspace{-10pt}\caption{Pairwise comparison of MMI-ALI with other baselines. Chance is at 50\%. Each cell indicates the percentage
			where our result is preferred over the other method. MMI-ALI overwhelmingly outperforms StarGAN and stay ahead of CycleGAN. }\label{t3}
		\begin{tabular}{lccccccccr}
			&\begin{tiny}Okapi2Zebra\end{tiny} &\begin{tiny}
				Okapi2Horse
			\end{tiny} &\begin{tiny}
			Zebra2Okapi
		\end{tiny} &\begin{tiny}
		Horse2Okapi
	\end{tiny} \\
	\hline
	StarGAN  &100.0\% &100.0\% &97.6\% & 100.0\%\\
	CycleGAN	& 57.2\%	&52.1\%	&56.5\%	&67.2\%			\\		
\end{tabular}\vspace{-10pt}
\end{footnotesize}\end{table}
\begin{figure}[h]\centering
	\vspace{-10pt}\includegraphics[width=3in]{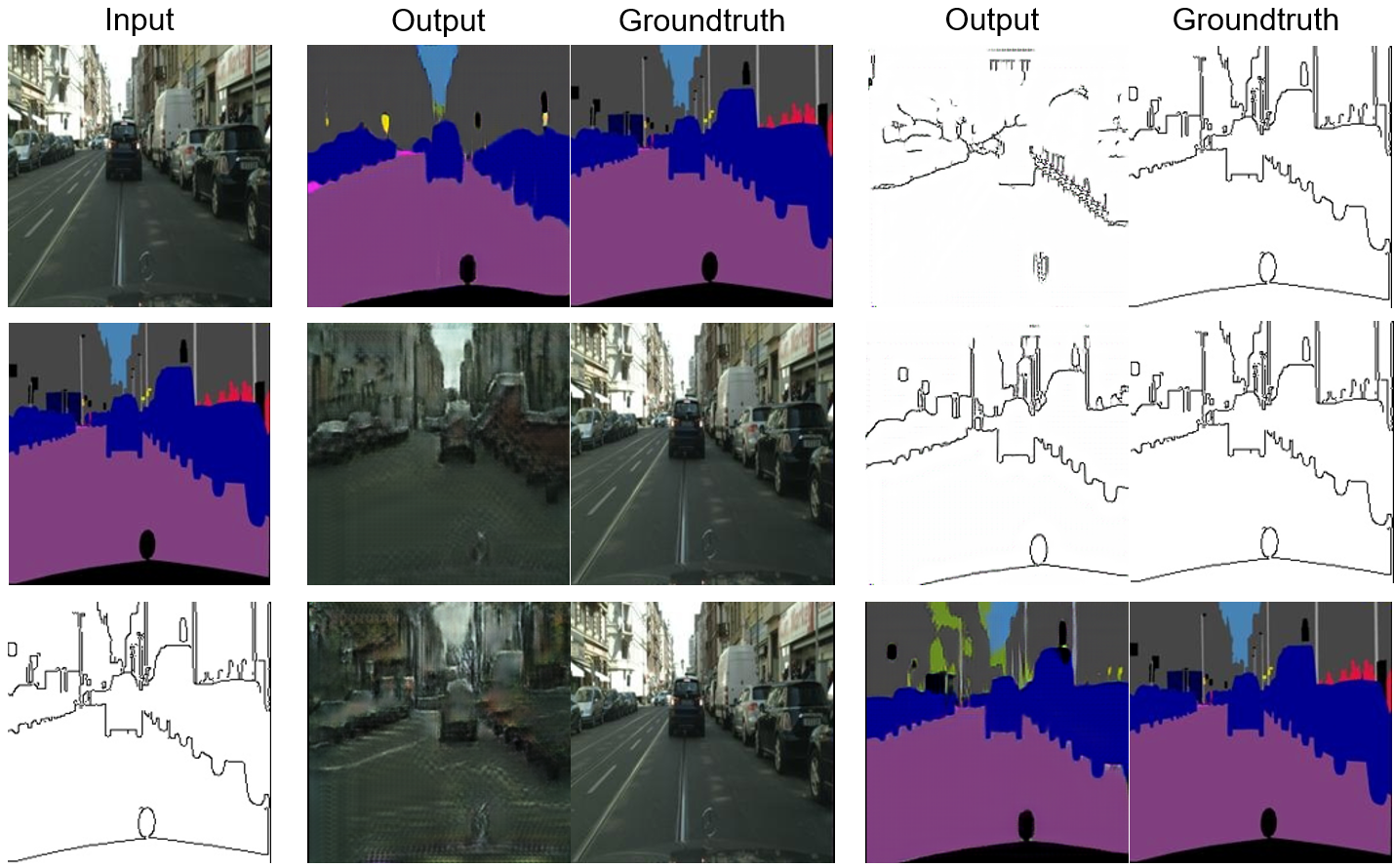}
	\caption{
		Cross-3-domain supervised transfer in Cityscape.  
	}\label{city1}\vspace{-8pt}
\end{figure}
\begin{figure}[t]\centering
	\includegraphics[width=3.2in]{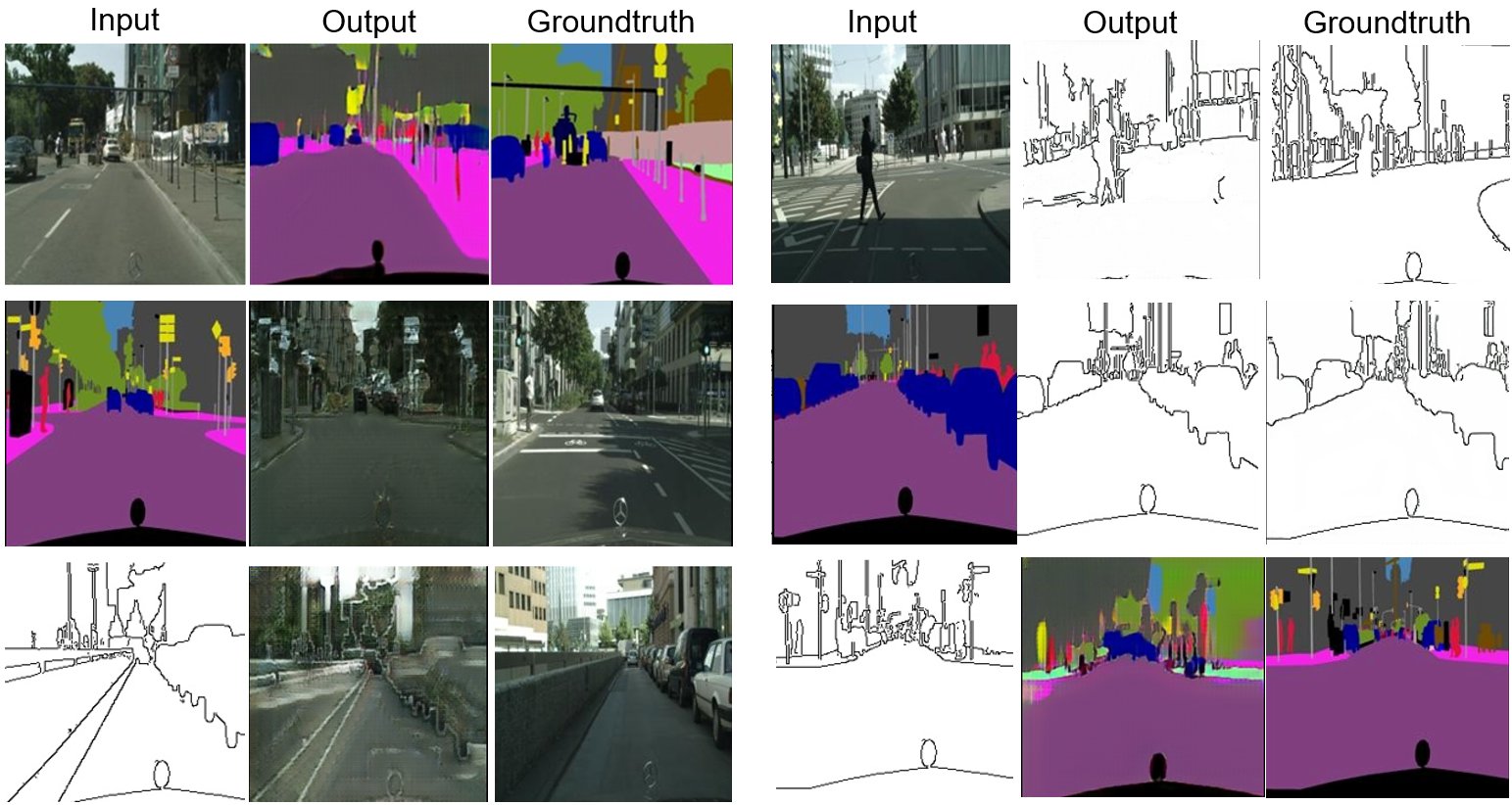}
	\vspace{-10pt}\caption{
		Cross-3-domain unsupervised transfer in Cityscape.  
	}\label{city2}\vspace{-16pt}
\end{figure}
\vspace{-2pt}

The visualization of object transfiguration are illustrated in Fig.\ref{vst}. First of all, StarGAN takes a mild effect. Due to the its category-generative pipeline, cross-domain style knowledge is hardly disentangled and thus, drives the produced images lack of fidelity in details. In a comparison, CycleGAN performs so aggressive that some details in the original images have been undesirably modified (Such negative effect is highlighted in red boxes). MMI-ALI 
successfully avoids the problem CycleGAN and StarGAN suffer from. Table.\ref{t3} shows the consistent quantitative results.

\begin{table}[h]
	\vspace{-16pt}\centering\caption{ FID in cross-$3$-domain transfer in Cityscape}\label{tFID}
	\fontsize{7.6}{9.4}{\selectfont
		\begin{tabular}{lccccccccr}
			\hline
			&&\begin{tiny}
				R$\hspace{-0.2em}\rightarrow\hspace{-0.2em}$Seg
			\end{tiny}\hspace{-0.2em} &\begin{tiny}Seg$\hspace{-0.2em}\rightarrow\hspace{-0.2em}$R\end{tiny}\hspace{-0.2em} &\begin{tiny}R$\hspace{-0.2em}\rightarrow\hspace{-0.2em}$Ske\end{tiny}\hspace{-0.2em} &\begin{tiny}
			Ske$\hspace{-0.2em}\rightarrow\hspace{-0.2em}$R
		\end{tiny}\hspace{-0.2em} &\begin{tiny}Seg$\hspace{-0.2em}\rightarrow\hspace{-0.2em}$Ske\end{tiny}\hspace{-0.2em} &\begin{tiny}Ske$\hspace{-0.2em}\rightarrow\hspace{-0.2em}$Seg\end{tiny}\\\hline
		\multirow{3}{*}{\rotatebox{90}{\begin{tiny}
					\textbf{Unsuper}
				\end{tiny}}}&ST	& 	405.16	&372.59		&385.08		& 388.97		&357.19		&417.39 			\\
				&CG	&224.04		&\textbf{213.43}		&164.65		&\textbf{222.24}	  &\textbf{60.20}	&\textbf{144.07}	\\
				&Ours&\textbf{202.93}		&254.41		&\textbf{150.98}		&246.04	  &101.30 &192.13\\
				\multirow{3}{*}{\rotatebox{90}{\begin{tiny}
							\textbf{Super}
						\end{tiny}}}&ST	&382.90 		&440.53		&419.11		&383.72		&400.70		&299.82 			\\
						&CG	&\textbf{217.28}		&260.41		&\textbf{171.04}		&\textbf{223.43}	  &65.18	&228.61	\\
						&Ours&250.48		&\textbf{246.01}		&196.06		&229.45	  &\textbf{55.76} &\textbf{143.20}\\
						\hline
					\end{tabular}
				}\vspace{-10pt}
			\end{table}
			In Cityscape, MMI-ALI achieved the leg-and-leg performances with CycleGAN in FID in supervised and unsupervised learning (Table \ref{tFID}). But CycleGAN gets less benefits from supervision. They significantly outperformed StarGAN. As observed in Fig \ref{city1} \ref{city2}, when MMI-ALI is compared with the target generation groundtruth, it has achieved superior transfers so that avoided modeling $C^2_{m}$ generators.      
			
			\vspace{-2pt}\subsection{Cross-$m$-emotion text style transfer.}\vspace{-4pt}
			In final experiment, we conduct a emotion style transfer in a text semantic embedding space. Specifically, we employ MojiTalk dataset \cite{Zhou2017MojiTalk} that contains $64$ emojis, and we collect a part of them to construct $4$ domains related to 'Happy' (40000 entries), 'Angry' (29000 entries), 'Pensive' (14000 entries) and 'Abash' (6261 entries), respectively. In this scenario, the goal of MMI-ALI is to transform the emotional text embeddings (we choose skip-thought \cite{Kiros2015Skip} as our language model to extract the representation of each text in the domains) from one domain to the others. 
			
\begin{figure}[t]\centering
			\vspace{-6pt}\includegraphics[width=3.2in]{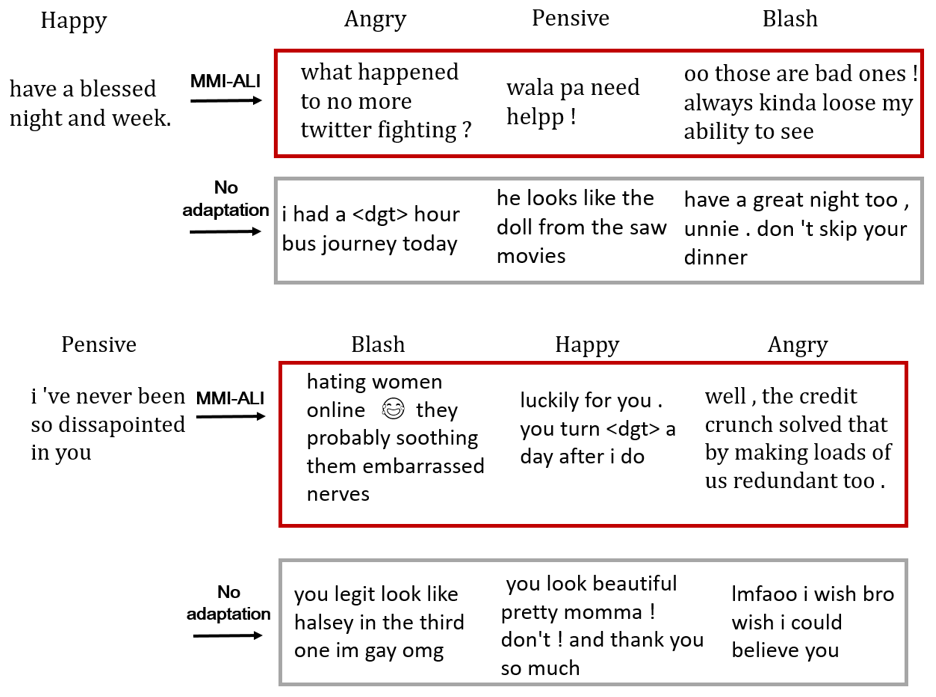}
			\caption{
				The illustration of emotion style transfer in skipthough embedding space. We compare our MMI-ALI with no adaptation.  
			}\vspace{-10pt}\label{nlp}
		\end{figure}

			\textbf{Evaluation.} Due to the embedding space is substantially discrete, the aforementioned metrics are not appropriate to evaluate the transfer efficiency. In this way, we employ a famous MRR (Mean Reciprocal Rank, \cite{Craswell2009Mean}), to measure the emotion transfer quality. For instance, when MMI-ALI transfer ``happy'' into ``angry'', we sort all sentences' embeddings based on their cosine distance to the embeddings generated from MMI-ALI. Then we calculate the rank of the nearest ``angry'' embedding and use its average of all transfer score. We use a simple fully-connected network with ReLU as the base backbone of MMI-ALI and train it with Batch normalization (BN). We compare MMI-ALI with the no-adaptation groundtruth results and the state-of-the-art unaligned text style transfer model \cite{shen2017style} that trained by the official code
			.The results are shown in Table.\ref{t4}. We provide more visualization by retrieving the nearest neighbor of each target domain, for the embeddings before (\emph{no adaptation}) and after MMI-ALI transform (Fig.\ref{nlp}). As can be observed, the transferred embeddings (outputs of MMI-ALI) leads to the neighbor embeddings with the texts containing more significant emotion.   
			\begin{table}[h]\center
				\begin{footnotesize}
					\vspace{-16pt}\caption{MRR for each domain transfer evaluation. Higher is better. As can be seen, MRRs in ``Happy'' and ``Abush'' are even higher than the original domain, indicating the effectiness of MMI-ALI.}\label{t4}. 
					\begin{tabular}{lccccccccr}
						&\begin{tiny}Happy\end{tiny} &\begin{tiny}
							Angry
						\end{tiny} &\begin{tiny}
						Pensive
					\end{tiny} &\begin{tiny}
					Abash
				\end{tiny} \\
				\hline
				groundtruth &0.71 &0.41 &0.53 & 0.21\\
				\cite{shen2017style} &0.52&0.17&0.31&0.07\\
				MMI-ALI	& 1.0	&0.40	&0.27	&0.24			\\		
			\end{tabular}\vspace{-10pt}
		\end{footnotesize}\end{table}  
		
		\vspace{-2pt}\section{Conclusion}\vspace{-4pt}
		In this paper, we have delved into the problem of multiple domain joint distribution matching that summarized a variety of cross-domain generation tasks.  Instead of hacking a complex DGM pipeline, we propose MMI-ALI, which reshapes classical ALI from the perspective of model integration and is linearly-scalable with the domain number. It learns with an adversarial ensemble loss and can be applied in both supervised and unsupervised learning schemes. Extensive evaluation results on diverse $m$-domain scenarios have demonstrated the superiority of the proposed framework to the existing DGMs feasible for cross-$m$-domain generation, e.g., CycleGAN and Star-GAN.
		\clearpage
		\section*{Acknowledgement} \begin{small}
			This work was supported in part by the National Key Research and Development Program of China under Grant No. 2018YFC0830103 and Grant No.2016YFB1001004, in part by National High Level Talents Special Support Plan (Ten Thousand Talents Program), and in part by National Natural Science Foundation of China (NSFC) under Grant No. 61622214,  61836012, and 61876224. Also, we thank Pengxu Wei
			for her valuable comments of this manuscript.
		\end{small}

\nocite{langley00}

\bibliography{example_paper}
\bibliographystyle{icml2019}

\appendix

\end{document}